%% file: 0.Main.tex
\documentclass{article}
\usepackage{ijcai25}

\usepackage{times}
\usepackage{soul}
\usepackage{url}
\usepackage[hidelinks]{hyperref}
\usepackage[utf8]{inputenc}
\usepackage[small]{caption}
\usepackage{graphicx}
\usepackage{amsmath}
\usepackage{amsthm}
\usepackage{booktabs}
\usepackage{algorithm}
\usepackage{algorithmic}
\usepackage[switch]{lineno}
\usepackage{tcolorbox}
\usepackage{diagbox}
\usepackage{multirow}
\usepackage{amsfonts}
\usepackage{stmaryrd}
\usepackage{savesym}
\usepackage{makecell}
\usepackage{xcolor}   % For colors
\usepackage{colortbl} % For coloring cells

\definecolor{customblue}{RGB}{176,209,228}
\title{$\text{M}^{2}$LLM: Multi-view Molecular Representation Learning\\with Large Language Models}

\author{Jiaxin Ju$^1$\and Yizhen Zheng$^2$\footnotemark[2]\and Huan Yee Koh$^{2,3}$\and Can Wang$^1$\And Shirui Pan$^1$\footnotemark[2]\\
\affiliations
$^1$School of Information and Communication Technology, Griffith University\\
$^2$Department of Data Science and AI, Monash University\\
$^3$Drug Discovery Biology, Monash Institute of Pharmaceutical Sciences, Monash University\\
\emails
jiaxin.ju@griffithuni.edu.au,
\{yizhen.zheng,huan.koh\}@monash.edu,
\{can.wang,s.pan\}@griffith.edu.au
}

\begin{document}

\maketitle

\begin{abstract}
    Accurate molecular property prediction is a critical challenge with wide-ranging applications in chemistry, materials science, and drug discovery. Molecular representation methods, including fingerprints and graph neural networks (GNNs), achieve state-of-the-art results by effectively deriving features from molecular structures. However, these methods often overlook decades of accumulated semantic and contextual knowledge. Recent advancements in large language models (LLMs) demonstrate remarkable reasoning abilities and prior knowledge across scientific domains, leading us to hypothesize that LLMs can generate rich molecular representations when guided to reason in multiple perspectives. To address these gaps, we propose $\text{M}^{2}$LLM, a multi-view framework that integrates three perspectives: the molecular structure view, the molecular task view, and the molecular rules view. These views are fused dynamically to adapt to task requirements, and experiments demonstrate that $\text{M}^{2}$LLM achieves state-of-the-art performance on multiple benchmarks across classification and regression tasks. Moreover, we demonstrate that representation derived from LLM achieves exceptional performance by leveraging two core functionalities: the generation of molecular embeddings through their encoding capabilities and the curation of molecular features through advanced reasoning processes.
\end{abstract}

\renewcommand\thefootnote{}\footnotetext{\textsuperscript{\dag}~Corresponding author.}

\input{Main_Text/1.Introduction.tex}
\input{Main_Text/2.Related_Work}
\input{Main_Text/3.Method}

\input{Main_Text/4.Experiment_Results}

\input{Main_Text/5.Conclusion}
\section*{Acknowledgments}
This research was partly funded by Australian Research Council (ARC) under grants FT210100097 and DP240101547.

%% The file named.bst is a bibliography style file for BibTeX 0.99c
\bibliographystyle{named}
\bibliography{m2llm}

\input{Main_Text/Appendix}

\end{document}

%% file: Main_Text/1.Introduction.tex
\section{Introduction}
\begin{figure}[t]
    \centering
    \includegraphics[width=0.47\textwidth]{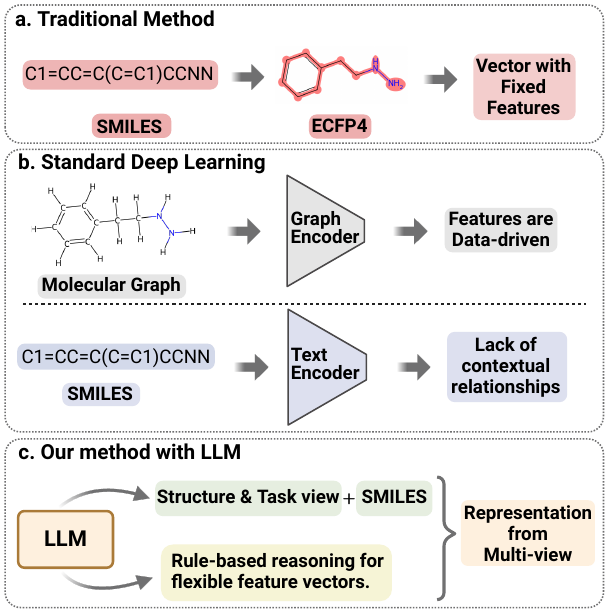}\\
    \caption{(a) Traditional Method: Converts SMILES into fixed ECFP4 vectors. (b) Standard Deep Learning: Graph encoders learn patterns from data, while text encoders for SMILES-only input lack contextual relationships. (c) $\text{M}^{2}$LLM: Generating multi-view representations by leveraging two core capabilities of LLMs: encoding contextual relationships and rule-based reasoning for molecular feature generation.}
    \label{fig.gap}
\end{figure}

Molecular property prediction is vital in cheminformatics~\cite{yang2019analyzing} and drug discovery~\cite{drews2000drug}, enabling the estimation of key characteristics like blood-brain barrier permeability, solubility, and toxicity. Traditional approaches rely heavily on predefined molecular descriptors, such as Extended-Connectivity Fingerprints (ECFPs)~\cite{rogers2010extended}, derived from SMILES(Simplified Molecular Input Line Entry System)~\cite{weininger1988smiles}. While these methods are efficient, their fixed feature sets limit their ability to capture the complex relationships needed for specific chemical tasks. Graph Neural Networks (GNNs) have demonstrated strong capabilities across a wide range of domains~\cite{bu2024improving,yu2024kernel,wang2024goodat,wang2024contrastive} and have also shown effectiveness in capturing structural and physicochemical patterns from molecular graphs~\cite{koh2024physicochemical,yu2025collaborative,du2024mmgnn}. However, they rely heavily on dataset-driven learning, which limits their ability to generalize across diverse chemical tasks. Similarly, recent studies~\cite{sadeghi2024comparative,shirasuna2024multi} leverage text encoder~\cite{medsker2001recurrent} or Transformer-based language model~\cite{kenton2019bert} using SMILES representations, which focus solely on string patterns and lack the ability to capture contextual relationships or molecular semantics.

To address these limitations, large language models (LLMs) offer a transformative approach to molecular representation by leveraging semantic and contextual knowledge from diverse pretraining corpora, including scientific literature and domain-specific datasets~\cite{zheng2024large}. In addition, LLMs demonstrate emergent abilities such as contextual reasoning, relational understanding, and the ability to extrapolate patterns, making them uniquely suited for tasks that require deep semantic understanding~\cite{kojima2022large,zheng2023large}. These capabilities far surpass those of earlier language models or text encoders. While those methods demonstrate the utility of LLMs in capturing molecular representations, they are inherently constrained by their exclusive reliance on SMILES as input. As a result, the potential of LLMs to enhance molecular representation by leveraging their pretrained scientific knowledge and emergent abilities remains underexplored.

In this work, we propose $\text{M}^{2}$LLM, a novel multi-view molecular representation learning framework that addresses these gaps by fully exploiting the power of LLMs. The framework is organized into two key modules: Molecular Embedding Generation and Molecular Feature Curation, each contributing distinct yet complementary perspectives on molecular data. The molecular embedding generation module leverages the semantic embedding ability of LLMs to represent molecular information from multiple views. This module includes the molecular structure view, which encodes structural information from SMILES sequences to capture structure-specific insights. This structure view can be further extended to incorporate additional insights, broadening the scope and enriching the molecular representation. Additionally, the molecular task view contextualizes molecules within specific prediction tasks to provide task-relevant guidance. Together, these views harness the LLM’s pretrained semantic knowledge to generate comprehensive embeddings. 

The molecular feature curation module, on the other hand, utilizes the reasoning ability of LLMs to derive interpretable features. This module introduces the molecular rules view, which generates rule-based features informed by scientific knowledge and observed data patterns, facilitating a deeper understanding of molecular properties. To unify these representations, $\text{M}^{2}$LLM employs a dynamic fusion mechanism that adaptively combines the contributions of each view based on the requirements of the task and the characteristics of individual molecules. The fused representation is then used for downstream prediction tasks, with a multi-layer perceptron (MLP) designed for both classification and regression. Through extensive experiments, we demonstrate that $\text{M}^{2}$LLM achieves state-of-the-art performance across multiple molecular property prediction benchmarks, highlighting the potential of LLMs to redefine molecular representation learning. 

Our contributions of this work are as follows: (1) We propose $\text{M}^{2}$LLM, a novel multi-view molecular representation learning framework that integrates diverse molecular perspectives through molecular structure view, molecular task view, and molecular rules view. These views are fused dynamically to create a unified representation tailored to each prediction task. (2) We explore the potential of representations derived from LLMs using $\text{M}^{2}$LLM for molecular property prediction, demonstrating that LLMs can achieve high performance by leveraging their dual capabilities: molecular embedding generation through their encoding abilities and molecular feature curation through their reasoning capabilities, addressing a significant gap in current research. (3) We show that $\text{M}^{2}$LLM achieves state-of-the-art performance across multiple molecular property prediction benchmarks, demonstrating its adaptability, scalability, and effectiveness in advancing the use of LLMs for molecular representation learning.

%% file: Main_Text/2.Related_Work.tex
\section{Related Work}
Several molecular representation methods have been developed, including molecular graph, ECFPs, and string line annotations such as SMILES. With the advancement of AI, machine learning models are now extensively used for property prediction through traditional and deep learning approaches.

In the conventional approach, traditional machine learning models, such as random forests~\cite{breiman2001random}, rely on computed molecular fingerprints to predict properties by capturing relationships between molecular substructures~\cite{jeon2019fp2vec}, though these predefined fingerprints may not fully capture complex molecular structural patterns and interactions. On the other hand, GNNs have been widely applied across domains~\cite{zhengonline,wu2024graph,zhang2025dynamic,ZhangMHLLL19}, they have also been effectively used to model molecular graphs~\cite{you2020graph,wang2022molecular,xia2022mole}, capturing hierarchical structural information and uncovering complex molecular patterns. However, these approaches often fail to integrate broader contextual, and may overlook knowledge already discovered in scientific literature and encoded within LLMs.

LLMs like GPT-4~\cite{achiam2023gpt} and Galactica~\cite{taylor2022galactica}, trained on diverse scientific and chemical datasets, capture semantic and contextual relationships beyond traditional text encoders' syntactic patterns. Moreover, LLMs exhibit emergent abilities such as reasoning~\cite{kojima2022large}, relational understanding~\cite{mirza2024large}, and pattern recognition~\cite{zheng2023large}, making them powerful tools for extracting insights from molecular text representations. Mirza et al.~\shortcite{mirza2024large} indicate that even a 7B-parameter LLM can achieve average human scores, while advanced models like GPT-4 can surpass the highest human scores in chemical reasoning.

Previous studies~\cite{wang2019smiles,fabian2020molecular,ross2022large} have explored encoding SMILES using LLMs as molecular embeddings, demonstrating their effectiveness in capturing meaningful representations and performance in downstream property prediction tasks~\cite{sadeghi2024comparative}. Researchers~\cite{luo2024learning,zhengcross,rollins2024molprop} have also investigated combining text encoders with GNNs to leverage both contextual and structural information. However, these methods remain limited by their reliance on SMILES as the sole input, failing to fully exploit the semantic depth of LLMs.

\begin{figure*}[t]
    \centering
    \includegraphics[width=0.95\textwidth]{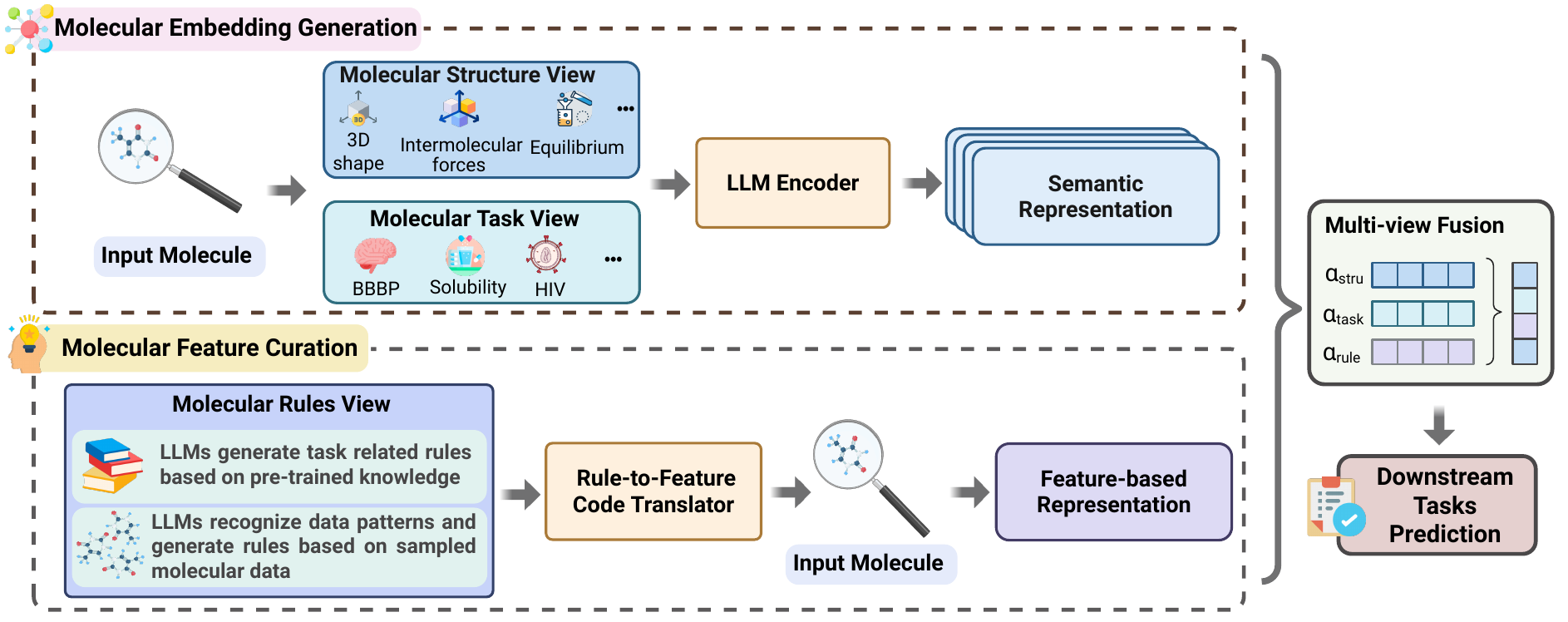}\\
    \caption{\textbf{The $\text{M}^{2}$LLM Molecular Representation Learning Framework.} The framework integrates three molecular views to create comprehensive and adaptable representations. Embedding generation includes two views: the molecular structure view and the molecular task view, both processed using LLMs to produce semantic representations. Feature curation provides one view: the molecular rules view, where rules are generated using LLMs based on pretrained knowledge and recognized data patterns, and translated into features via a rule-to-feature translator to produce feature-based representation. These diverse representations are fused in a multi-view fusion module, with trainable weights ($\alpha$) balancing each component, enabling accurate predictions through a MLP prediction.}
    \label{fig.framework}
\end{figure*}

%% file: Main_Text/3.Method.tex
\section{Multi-view Molecular Representation Learning with Large Language Models}\label{framework}

In this section, we introduce $\text{M}^{2}$LLM, a novel multi-view framework designed to leverage LLMs for molecular representation learning. As illustrated in Figure \ref{fig.framework}, the framework consists of two main components: Embedding Generation and Feature Curation, which collectively provide three distinct views of molecular information. These views are fused in a Multi-View Fusion Module, generating a unified representation optimized for downstream prediction tasks.

\subsection{Molecular Embedding Generation}
The Molecular Embedding Generation component leverages LLMs to encode sequence inputs, providing two views: the Molecular Structure View and the Molecular Task View.

\subsubsection{Molecular Structure View}
The molecular structure view uses LLMs to generate embeddings that capture key physical and chemical properties of molecules. LLMs have demonstrated strong capabilities in encoding semantic and contextual knowledge. These models generate outputs sequentially, relying on previous input tokens to reason about the current context. Instead of directly providing SMILES as input, we frame specific questions about the molecule alongside its SMILES representation. This approach gives the LLM additional contextual information, allowing it to ``think about" the molecule in relation to the queried property, resulting in richer and more meaningful representations. To achieve this, we define three example insights as questions targeting various structural aspects of the molecule, which can be adapted or replaced as needed.

\begin{tcolorbox}
[boxsep=0mm,left=2.5mm,right=2.5mm,colframe=black!55,colback=customblue,fontupper=\normalsize]
\textbf{\textit{Structure Insight 1:}} How does the molecule’s 3D shape change in different environments, and what are the effects of these changes?

\textbf{\textit{Structure Insight 2:}} What are the key intermolecular forces that govern the behavior of this molecule in various contexts?

\textbf{\textit{Sructure Insight 3:}} How does the molecule contribute to the overall chemical equilibrium in its different environments?
\end{tcolorbox}

%These insights are paired with the molecule’s SMILES representation to form structured inputs. 
Let $s_i$ denote the SMILES for molecule $i$, and $q_j$ denote one of the three structure insight questions. The structural embedding $z_{ij}^{\text{struct}}$ for molecule $i$ with question $q_j$ is computed as:

\begin{equation}
z_{ij}^{\text{struct}} = f_{\text{Encode}}^{\text{struct}}([q_j; s_i])
\end{equation}
where $f_{\text{Encode}}(\cdot)$ is the LLM encoder used to process the input, $[q_j; x_i]$ represents the concatenation of the question $q_j$ and the SMILES $s_i$. The embeddings generated for the three questions are then concatenated to form the final structure view representation:

\begin{equation}
z_i^{\text{struct}} = [z_{i1}^{\text{struct}}; z_{i2}^{\text{struct}}; z_{i3}^{\text{struct}}]
\end{equation}

Molecular structure view not only enriches the molecular structure representation but also provides a flexible framework for generating diverse and comprehensive embeddings. By incorporating structural views, $\text{M}^{2}$LLM focuses solely on analyzing molecular information without being tied to specific tasks or datasets. This design allows the LLM to encode the input in a way that leverages the semantic knowledge learned during the pretraining process, capturing more contextual and meaningful information than using SMILES alone. Additionally, the modular nature of the structural view facilitates straightforward extensions by introducing new questions to create additional views. 
%This adaptability ensures that the framework can be expanded to explore other aspects of molecular analysis, making it a scalable solution for diverse representation needs.

\subsubsection{Molecular Task View}
The molecular task view encodes task-specific information by framing molecular analysis as a natural language processing problem, inspired by the pre-training process of the Galactica model~\cite{taylor2022galactica} that has demonstrated the ability to process and reason about molecular representations effectively when supplemented with contextual information, such as questions or prompts. To leverage this capability, the task view combines a molecule's SMILES representation with a task-specific question, guiding the LLM to generate a task-aware representation by retrieving and processing semantic information encoded during pretraining. For example, consider the task of predicting blood-brain barrier penetration for a given molecule. The input to the LLM would take the following form:

\begin{tcolorbox}
[boxsep=0mm,left=2.5mm,right=2.5mm,colframe=black!55,colback=customblue,fontupper=\normalsize]
\textbf{Here is a SMILES formula:} [START\_I\_SMILES]C1=\\CC=C(C=C1)C(=O)O[END\_I\_SMILES]\\

\textbf{Question:} Will the chemical compound penetrate the blood-brain barrier?
\end{tcolorbox}

Unlike the molecular structure view, which focuses on general analysis of molecular properties, the task view explicitly tailors its input to the prediction problem at hand. The generated embedding $z_i^{\text{task}}$ incorporates both the SMILES sequence $s_i$ and the task-specific question $t$, and is computed as follows:

\begin{equation}
z_i^{\text{task}} = f_{\text{Encode}}^{\text{task}}([t; s_i])
\end{equation}

%Molecular task view takes advantage of the LLM's pretraining on diverse chemical datasets, enabling it to extract and encode task-relevant patterns from the SMILES input. Furthermore, using a question-driven approach aligns with the LLM's ability to process input sequentially, where the question provides critical context for interpreting the SMILES representation. The LLMs architecture and training approach, such as Galactica series, highlight the importance of combining structured domain-specific data, such as SMILES, with natural language context. By incorporating special tokens like [START\_I\_SMILES] and [END\_I\_SMILES] and posing the task as a question, the task view representation ensures that the LLM can leverage its pretrained knowledge to ``think through" the input and produce an embedding tailored to the specific prediction task. Molecular task view allows for adaptability and extensibility, as new tasks can be incorporated seamlessly by defining new task-specific questions.

\subsection{Molecular Feature Curation}
The Molecular Feature Curation component introduces an additional view of molecular representation through the Molecular Rules View. This view captures domain-specific knowledge and patterns by leveraging LLMs to generate rules based on pretrained knowledge and data-driven insights. These rules are then transformed into features using a Rule-to-Feature Code Translator, providing a numerical representation that complements the semantic representations generated by Molecular Embedding Generation module.

\subsubsection{Molecular Rules View}
The Molecular Rules View captures both pretrained knowledge from scientific literature and patterns derived from molecular datasets. The generated rules are transformed into numerical features, creating a representation that complements the semantic representations.

\paragraph{Scientific Rule Generation with LLMs:} LLM has built-in knowledge and an understanding of various tasks from its pre-training. To make use of this, we assign the LLM a specific persona, such as an experienced chemist, and instruct it to generate rules based on its extensive exposure to scientific literature. These rules are generated independently of any specific molecule and are instead tailored to the requirements of a given task. By leveraging its pretraining on vast scientific datasets, the LLM produces rules that reflect well-established principles and patterns relevant to the task at hand. Let $t$ represent the specific task, the generated rules, $R_{\text{sci}}$, for task $t$ can be expressed as $R_{\text{sci}}(t)=f_{\text{Reason}}^{\text{sci}}(t)$, where $f_{\text{Reason}}^{\text{sci}}(\cdot)$ leverages the LLM’s pretrained reasoning capabilities to derive rules. The example below demonstrates how the LLM generates scientifically grounded rules for predicting blood-brain barrier penetration in Scientific Rule Generation phase.
\begin{tcolorbox}
[boxsep=0mm,left=2.5mm,right=2.5mm,colframe=black!55,colback=customblue,fontupper=\normalsize]
\textbf{Persona}: Assume you are an experienced Chemist. Please come up with 20 rules that are important to predict if a molecule can penetrate the blood-brain barrier.\\

\textbf{LLM Answer Example}: \\
Rule 1: Molecular weight $<$ 500 Da\\
Rule 2: LogP value between 1 and 3\\
Rule 3: Presence of aromatic rings\\
...
\end{tcolorbox}
\paragraph{Data Pattern Rule Observation with LLMs:} Beyond their extensive knowledge base, LLMs exhibit strong abilities in identifying patterns and relationships~\cite{zheng2023large}, making them well-suited for recognizing task-relevant trends within molecular data. In this phase, several randomly selected subsets of SMILES strings in the training data $\{s_i\}_{i=1}^m \in S_{\text{train}}$ with their corresponding label $y_i$ are then provided to the LLM. By analyzing these subsets, additional rules $R_{\text{data}}$ based on observed patterns and relationships within the molecular structures for the specific task $t$:
$R_{\text{data}}(t) = \bigcup_{k=1}^K f_{\text{Reason}}^{\text{data}}(\{s_i, y_i\}_{i=1}^m, t)_k$,
where $K$ is the number of subsets analyzed, $m$ is the number of molecule in each subset, and $f_{\text{Reason}}^{\text{data}}(\cdot)$ leverages the LLM’s emergent reasoning capabilities to identify patterns and generate rules. The following example illustrates how the LLM generates task-specific rules by analyzing randomly selected one subset of molecular training data paired with their corresponding labels.
\begin{tcolorbox}
[boxsep=0mm,left=2.5mm,right=2.5mm,colframe=black!55,colback=customblue,fontupper=\normalsize]
\textbf{Persona}: Assume you are a very experienced Chemist. In the following data, with label 1, it means the smiles string is BBBP. With label 0, it means the smiles string is not BBBP. Please infer step-by-step to come up with 3 rules that directly relate the properties/structures of a molecule to predict if it can be BBBP. \\
\text{[}\textit{\textbf{SMILES strings along with corresponding labels}}\text{]}\\

\textbf{LLM Answer Example}: \\
Rule 1: The presence of a benzene ring in the molecule is essential for predicting whether it can be BBBP. \\
Rule 2: The presence of a carbonyl group (-C=O) in the molecule is also important for predicting whether it can be BBBP. \\
Rule 3: Number of Hydrogen Bond Donors (HBD)
\end{tcolorbox}
\subsubsection{Feature-based Representation}
The combined set of rules, $R(t)=R_{\text{sci}}(t)\bigcup R_{\text{data}}(t)$, is transformed into numerical features through a Rule-to-Feature Code Translator, which maps the rules, $\{r_1, r_2,\dots, r_n\} \in R(t)$, into a feature vector for each SMILES string $s$. This translation process leverages LLMs to transform text-based rules into code-based features and can be defined as:

\begin{equation}
f_i(s) = 
\begin{cases} 
1, & \text{if } r_i \text{ is satisfied by } s \\ 
0, & \text{otherwise,}
\end{cases}
\end{equation}
or alternatively as a numerical function $f_i(s) \in \mathbb{R}^n$ if the rule outputs a continuous value (e.g., molecular weight). The LLM is tasked with translating these rules into executable functions $f_i(\cdot)$, which are applied to the input $s$ to extract a set of molecular features. This process generates a feature vector $z_i^{\text{rule}}$ for a molecule $s$, where each feature corresponds to a rule:

\begin{equation}
z_i^{\text{rule}} = [f_1(s), f_2(s), \dots, f_n(s)]
\end{equation}

\subsection{Multi-view Representation Fusion}
The $\text{M}^{2}$LLM framework then integrates three views into a unified representation through a fusion mechanism. %This integration ensures that information from diverse perspectives is utilized effectively, enabling the framework to generate robust predictions tailored to specific molecular property prediction tasks. 
The fused representation is subsequently passed through a prediction module, which adapts to the requirements of either classification or regression tasks.

For a given molecule $s_i$, let $z_i^{\text{struct}}$, $z_i^{\text{task}}$, and $z_i^{\text{rule}}$ denote the representations obtained from the structure view, task view, and rules view, respectively. %These representations capture distinct yet complementary aspects of the molecule, such as structural features, task-specific contexts, and rules derived from scientific and data-driven patterns. 
To combine these representations into a single, comprehensive vector, our proposed framework employs a weighted sum mechanism. Each view's contribution is modulated by a set of learnable weights $\alpha_i^{\text{struct}} $, $\alpha_i^{\text{task}}$, and $\alpha_i^{\text{rule}}$, which are specific to each molecule. The fused representation $z_i^{\text{fused}}$ is then computed as a weighted sum of the individual view representations:

\begin{equation}
z_i^{\text{fused}} = \alpha_i^{\text{struct}} z_i^{\text{struct}} + \alpha_i^{\text{task}} z_i^{\text{task}} + \alpha_i^{\text{rule}} z_i^{\text{rule}}
\end{equation}
where weights satisfy: $\alpha_i^{\text{struct}} + \alpha_i^{\text{task}} + \alpha_i^{\text{rule}} = 1$, $\alpha_i^{\text{struct}}, \alpha_i^{\text{task}}, \alpha_i^{\text{rule}} \geq 0$. %This ensures that the contributions of each view are dynamically adjusted based on the characteristics of the individual molecule and task. 
The fused representation $z_i^{\text{fused}}$ is then used as input to a multi-layer perceptron (MLP), which performs the final prediction. %For classification tasks, the MLP outputs a probability distribution over the possible classes, while for regression tasks, it predicts a continuous value. 
This process is mathematically defined as:
\begin{equation}
\hat{y}_i = f_{\text{MLP}}(z_i^{\text{fused}})
\end{equation}
where $\hat{y}_i$ is the predicted output for molecule $s_i$, and $f_\text{MLP}(\cdot)$ represents the function of the multi-layer perceptron. 

The framework is trained to optimize the weights $\alpha_i^{\text{struct}} $, $\alpha_i^{\text{task}}$, and $\alpha_i^{\text{rule}}$, and the MLP parameters using task-specific loss functions. For classification tasks, the cross-entropy loss is minimized:
\begin{equation}
\mathcal{L}_{\text{cls}} = -\frac{1}{N} \sum_{i=1}^N \left[ y_i \log \hat{y}_i + (1 - y_i) \log (1 - \hat{y}_i) \right]
\end{equation}
For regression tasks, the root mean squared error (RMSE) loss is used:
\begin{equation}
\mathcal{L}_{\text{regression}} = \sqrt{\frac{1}{N} \sum_{i=1}^N (y_i - \hat{y}_i)^2}
\end{equation}
where $N$ denotes the number of molecules, $y_i$ represents the binary ground truth label or the true value, and $\hat{y}_i$ is the predicted probability or the predicted value.

%% file: Main_Text/4.Experiment_Results.tex
\begin{table*}[!htbp]
\centering
\resizebox{0.98\textwidth}{!}{%
\begin{tabular}{cc|c|c|c|c|c|c|c|c}
    \toprule
    \toprule
     & \diagbox{\textbf{Model}}{\textbf{Dataset}} & \makecell{\textbf{Backbone} \\ \textbf{Type}} & \textbf{BBBP(1) ↑} & \textbf{BACE(1) ↑} & \textbf{ClinTox(1) ↑} & \textbf{HIV(1) ↑} & \textbf{SIDER(27) ↑}& \textbf{Average} & \makecell{\textbf{Average} \\ \textbf{Rank}}\\ 
    \midrule
    \multirow{10}{*}{\rotatebox[origin=c]{90}{\textbf{Baselines}}} 
    & \textbf{RF + ECFP4} & RF & 67.6 ± 1.0 & \colorbox{orange!40}{\underline{\textbf{85.0 ± 1.2}}} & 69.4 ± 3.1 & 77.1 ± 0.7 & 62.6 ± 2.5 & 72.3 ± 1.7& 7\\ 
    \cmidrule(lr){2-8}
    & \textbf{AttrMask}~\cite{hu2019strategies} & GNN & 65.2 ± 1.4 & 77.8 ± 1.8 & 73.5 ± 4.3 & 75.3 ± 1.5 & 55.7 ± 4.0 & 69.5 ± 2.6 & 10\\ 
    & \textbf{GraphCL}~\cite{you2020graph} & GNN & 67.8 ± 2.4 & 74.6 ± 2.1 & 77.5 ± 3.4 & 75.1 ± 0.7 & 53.1 ± 4.3 & 69.6 ± 2.6& 9\\ 
    & \textbf{GraphMVP}~\cite{liu2022pre} & GNN & 70.8 ± 0.5 & 79.3 ± 1.5 & 79.1 ± 2.8 & 76.0 ± 0.1 & 59.6 ± 3.9 & 73.0 ± 1.8 & 6\\ 
    & \textbf{3D-infomax}~\cite{stark20223d} & GNN & 69.1 ± 1.2 & 78.6 ± 1.9 & 62.7 ± 3.3 & 76.1 ± 1.3  & 60.7 ± 3.1 & 69.4 ± 2.2 & 11\\
    & \textbf{MolCLR}~\cite{wang2022molecular} & GNN & 73.1 ± 1.6 & 81.5 ± 1.6 & 91.6 ± 2.7 & 77.3 ± 1.3 & 60.0 ± 2.9 & 76.7 ± 2.0 & 5\\ 
    & \textbf{MoleBert}~\cite{xia2022mole} & GNN & 71.9 ± 1.6 & 80.8 ± 1.4 & 78.9 ± 3.0 & 78.2 ± 0.8 & 51.4 ± 5.0 & 72.2 ± 2.4 & 8\\ 
    & \textbf{Uni-Mol}~\cite{zhou2023unimol} & Transformer & 71.5 ± 1.4 & \colorbox{orange!20}{\textbf{84.4 ± 2.1}} & 87.8 ± 2.6 & 78.3 ± 1.3 & 62.3 ± 5.6 & 76.9 ± 2.6 & 4\\  
    & \textbf{GROVER}~\cite{rong2020self} & Transformer & 65.1 ± 2.5 & 81.1 ± 2.3 & 74.0 ± 11.8 & 57.7 ± 4.3 & 56.8 ± 5.1 & 67.0 ± 5.2& 12\\ 
    %& \textbf{LLM4SD}~\cite{zheng2023large} & LLM & 74.5 ± 0.2 & 83.8 ± 0.3 & 92.8 ± 0.5 & \colorbox{orange!20}{\textbf{79.0 ± 0.2}} & \colorbox{orange!40}{\underline{\textbf{65.4 ± 1.4}}} & \colorbox{orange!20}{\textbf{79.1 ± 0.5}} & \textbf{2}\\ 
    \cmidrule(lr){2-8}
    \multirow{3}{*}{\rotatebox[origin=c]{90}{\textbf{Ours}}}
    & \textbf{$\text{M}^{2}$LLM($LLaMa$-$3.1$)} & LLM & \colorbox{orange!40}{\underline{\textbf{77.0 ± 1.0}}} & 77.8 ± 2.9 & 99.1 ± 0.4 & 77.2 ± 0.9 & 62.7 ± 0.4 & 78.8 ± 1.1 & 3\\
    & \textbf{$\text{M}^{2}$LLM($Galactica$)} & LLM & 74.9 ± 0.74 & 80.0 ± 2.7 & \colorbox{orange!20}{\textbf{99.4 ± 0.1}} & \colorbox{orange!20}{\textbf{77.5 ± 0.7}} & \colorbox{orange!20}{\textbf{62.8 ± 0.4}} & \colorbox{orange!20}{\textbf{79.0 ± 0.9}} & \textbf{2} \\
    & \textbf{$\text{M}^{2}$LLM($OpenAI$)} & LLM & \colorbox{orange!20}{\textbf{75.5 ± 1.3}} & 78.2 ± 0.9 & \colorbox{orange!40}{\underline{\textbf{99.5 ± 0.1}}} & \colorbox{orange!40}{\underline{\textbf{79.5 ± 0.7}}} & \colorbox{orange!40}{\textbf{63.7 ± 0.3}} & \colorbox{orange!40}{\textbf{79.3 ± 0.7}} & \textbf{1}\\ 
    \bottomrule
    \bottomrule
\end{tabular}}
\caption{Results on molecular property classification tasks with scaffold split. Mean and standard deviation of ROC-AUC (\%) from 10 random seeds are reported, with higher values indicate better performance. The top-2 performances on each dataset are shown in bold, with
\colorbox{orange!40}{\underline{\textbf{bold}}} being the best result, and \colorbox{orange!20}{\textbf{bold}} being the second best result.}
\label{class_table}
\end{table*}

\section{Experiment}
\subsection{Experimental Setup}
\paragraph{Dataset} Our framework is evaluated on 8 datasets spanning 34 tasks from MoleculeNet~\cite{wu2018moleculenet}, including physiology-related tasks like BBBP~\cite{martins2012bayesian}, ClinTox~\cite{gayvert2016data}, and 27 SIDER tasks~\cite{kuhn2016sider} for adverse drug reaction prediction. Additionally, we evaluate classification tasks from BACE~\cite{subramanian2016computational} and HIV~\cite{wu2018moleculenet}, as well as regression tasks from ESOL~\cite{delaney2004esol}, FreeSolv~\cite{mobley2014freesolv}, and Lipophilicity~\cite{wu2018moleculenet}. We use the scaffold splitting method recommended by MoleculeNet~\cite{wu2018moleculenet}, which assigns molecules with distinct structural scaffolds to separate training, validation, and test sets. This method, unlike random splitting, ensures structural dissimilarity between sets, creating a more challenging evaluation scenario. Detailed dataset descriptions are provided in Appendix A.1.%\ref{Adx:datasets}.

\paragraph{Baselines} For the traditional model, we use Random Forest~\cite{breiman2001random} with ECFP4~\cite{rogers2010extended} as the input feature set. For deep learning models, as shown in Table \ref{class_table}, we select the most representative GNNs pretraining baselines and transformer-based architecture. To ensure a fair comparison, We rerun all baseline models with the same random seed across 10 iterations.

\paragraph{Backbone Model} Our framework is built on an LLM-based multi-view architecture, utilizing a state-of-the-art LLM as its backbone. Specifically, for the molecular structural and task views, we use Galactica models (6.7B and 30B parameters)~\cite{taylor2022galactica}, LLaMa-3.1 models (8B and 8B-instruct)~\cite{dubey2024llama}, and OpenAI's closed-source text embedding models (small and large configurations)~\cite{openai2024embedding}. For the molecular rules view, rule generation is performed using the Galactica models, leveraging their extensive pretraining on scientific literature to produce high-quality task-specific rules.

\subsection{Performance on Classification Tasks}
\label{cls_performance}
We evaluate $\text{M}^{2}$LLM on five classification datasets with 31 subtasks, as shown in Table \ref{class_table}. We report the mean and standard deviation from 10 random seeds using the evaluation metric, receiver operating characteristic-area under the curve (ROC-AUC) (\%), where higher scores indicate better performance. One result for the same LLM backbone architecture is presented for the comparison with other state-of-the-art baselines. Full results for different backbone architectures are provided in the Appendix A.2.%\ref{diff_backbone}.

As shown in Table \ref{class_table}, our framework demonstrates superior performance, surpassing existing baselines with significant improvements. Notably, our framework exhibits exceptional performance on the Clintox dataset, achieving a near-perfect accuracy of 99.5\% and 99.4\%, this result significantly outperforms all other models. Moreover, on the BBBP, HIV, and SIDER datset, our framework variants achieve the best and second-best results, outperforming all GNN-based and Transformer-based baselines. This result further enhances the credibility of LLM-based approaches in molecular property prediction tasks. Full results on 27 tasks for the SIDER Dataset can be found in Appendix A.3. In the case of the BACE dataset, $\text{M}^{2}$LLM achieves 80.0\%, which, while competitive, remains below the highest baseline result of 85.0\% achieved by the RF model. The potential reasons may be the BACE dataset assigns binary labels for molecular inhibitors of human $\beta$-secretase 1 (BACE-1), based on an arbitrary threshold of quantitative potency values (IC\textsubscript{50}) set at 7~\cite{wu2018moleculenet}. However, potency values can vary significantly depending on the assay settings~\cite{landrum2024combining}, lower potency values can still indicate strong inhibition of BACE-1~\cite{harding2024iuphar}. We hypothesize that this arbitrary threshold and label ambiguity hinder LLMs' ability to reason effectively.

\subsection{Performance on Regression Tasks}
We evaluate $\text{M}^{2}$LLM on three regression tasks, as shown in Table \ref{reg_table}. We report the RMSE for regression, where lower values signify better result. Results presented in Table \ref{reg_table} demonstrate the superior performance of our proposed framework compared to all baselines across three datasets. Specifically, $\text{M}^{2}$LLM demonstrates strong performance, achieving an RMSE of 2.01 on FreeSolv dataset, reducing the error by 15.5\% compared to the best baseline value of 2.38. Furthermore, on ESOL dataset, it achieves an RMSE of 0.44, a remarkable 56.9\% reduction in error compared to the best baseline value of 1.02. Additionally, on the Lipophilicity dataset, it achieves state-of-the-art results with an RMSE of 0.66, while our other variants demonstrate competitive performance against baseline models. 

\begin{table}[!ht]
\centering
\resizebox{0.49\textwidth}{!}{%
\begin{tabular}{c|c|c|c}
\toprule
\toprule
\diagbox{\textbf{Model}}{\textbf{Dataset}} & \textbf{ESOL(1) ↓} & \textbf{FreeSolv(1) ↓} & \textbf{Lipophilicity(1) ↓}\\  
\midrule
    \textbf{RF + ECFP4} & 1.34 ± 0.01 & 4.36 ± 0.04 & 0.90 ± 0.00\\ 
    \cmidrule(lr){1-4}
    \textbf{AttrMask} & 1.11 ± 0.05 & 2.92 ± 0.03 & 0.73 ± 0.00 \\ 
    \textbf{GraphCL} & 1.31 ± 0.07 & 3.60 ± 0.32 & 0.78 ± 0.02\\ 
    \textbf{GraphMVP} & 1.06 ± 0.02 & 2.95 ± 0.19 & 0.69 ± 0.01 \\ 
    \textbf{3D-infomax} & 0.89 ± 0.04 & 2.83 ± 0.10 & 0.70 ± 0.02 \\
    \textbf{MolCLR} & 1.31 ± 0.03 & 2.73 ± 0.08 & 0.74 ± 0.02 \\ 
    \textbf{MoleBert} & 1.02 ± 0.03 & 3.08 ± 0.05 & \colorbox{orange!20}{\textbf{0.68 ± 0.02}} \\ 
    \textbf{Uni-Mol} & 1.55 ± 0.26 & 3.94 ± 0.50 & 1.19 ± 0.07 \\  
    \textbf{GROVER} & 1.13 ± 0.08 & 2.38 ± 0.40 & 0.91 ± 0.09\\ 
    %\textbf{LLM4SD} & 0.52 ± 0.04 & 2.62 ± 0.01 & \colorbox{orange!20}{\textbf{0.68 ± 0.00}}\\ 
\cmidrule(lr){1-4}
    \textbf{$\text{M}^{2}$LLM($LLaMa$-$3.1$)} & \colorbox{orange!40}{\underline{\textbf{0.44 ± 0.01}}} & \colorbox{orange!40}{\underline{\textbf{2.01 ± 0.37}}} & 0.73 ± 0.03\\
    \textbf{$\text{M}^{2}$LLM($Galactica$)} & 0.53 ± 0.25& 2.39 ± 1.39 &\colorbox{orange!40}{\underline{\textbf{0.66 ± 0.02}}}\\
    \textbf{$\text{M}^{2}$LLM($OpenAI$)} & \colorbox{orange!20}{\textbf{0.48 ± 0.02}} & \colorbox{orange!20}{\textbf{2.35 ± 0.63}} & 0.77 ± 0.01\\
\bottomrule
\bottomrule
\end{tabular}}
\caption{Results on Molecular Property Regression tasks with Scaffold Split. Mean and standard deviation of the Root Mean Square Error (RMSE) metric from 10 random seeds are reported, with lower scores indicating better performance. Average statistics of target labels are -3.46 for ESOL, -6.33 for FreeSolv, and 2.20 for Lipophilicity.}
\label{reg_table}
\end{table}

\subsection{Multi-view Component Contribution Analysis}
\label{comp_contri}

\begin{figure}[!ht]
    \centering
       \includegraphics[width=0.46\textwidth]{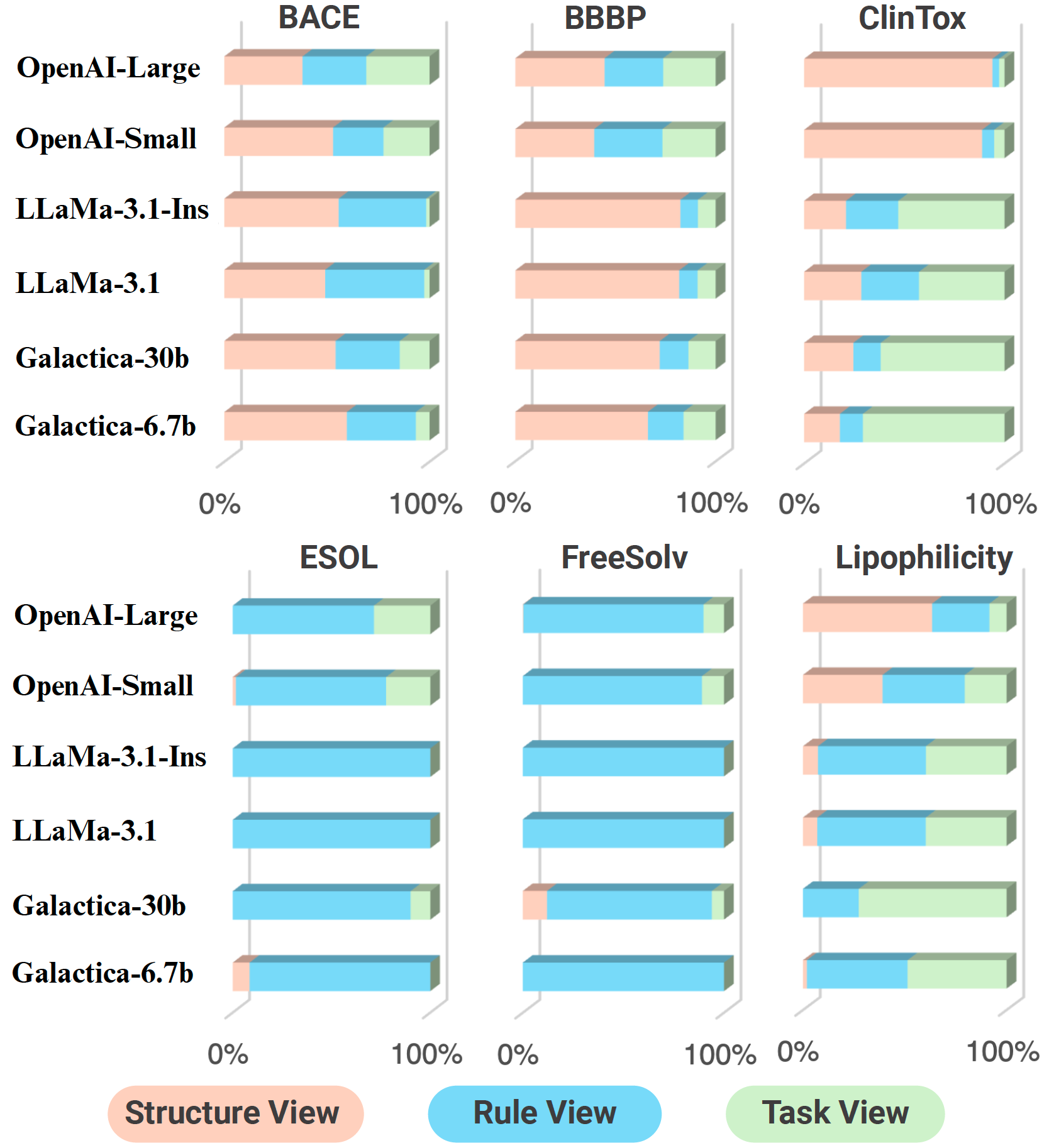}\\
       \caption{Multi-view Component Contribution Analysis. The contributions of the Molecular Structure View, Molecular Rules View, and Molecular Task View to the final score are calculated by averaging component weights for each SMILES representation across 10 random seeds.}
       \label{fig.compontent}
\end{figure}

In this section, we analyze the contribution of each view component to the final decision, as illustrated in Figure \ref{fig.compontent}, based on three classification datasets and three regression tasks. Interestingly, the molecular structure view component contributes more significantly to the classification tasks, whereas the molecular rule view component and molecular task view component play a larger role in the regression tasks. This suggests that classification tasks may benefit from a detailed representation of molecular structures, as these tasks often rely on recognizing specific structural features critical for distinguishing between categories. On the other hand, regression tasks, which predict continuous values such as molecular properties, appear to benefit more from the molecular rule and task views, which capture broader context and relationships. This demonstrates that our framework effectively automates the learning of appropriate weights for each component, dynamically optimizing the contribution of each view for individual molecules based on the task requirements.

From the LLM architecture perspective, we find that LLMs with the same architecture tend to exhibit similar component contributions across different datasets to the final predictions. However, on the ESOL and FreeSolv datasets, even though the LLMs heavily rely on one or two components, Galactica-6.7B and 30B demonstrate different behavior, despite having the same architectural design, these models exhibit different component contributions for their final decisions. This pattern highlights the flexibility and adaptability of our proposed multi-view representation learning framework.
 
Furthermore, the results for the ClinTox dataset, as reported in Table \ref{class_table}, demonstrate that we achieve near-perfect scores across all backbone settings. However, this component analysis reveals intriguing insights into how different model architectures rely on various components to make their predictions. The OpenAI models heavily depend on the molecular structure view, indicating that their decision-making process is primarily driven by structural understanding and extensive pre-trained knowledge. In contrast, the Galactica models rely more on task-specific components, likely because our task-specific thinking process is closely aligned with their pre-training dataset and methodology. The LLaMa-3.1 models demonstrate a relatively balanced utilization of all three components to make accurate predictions.

\subsection{Effectiveness of Multi-view Representation}
\label{rep_effective}
To better understand the performance gains afforded by our proposed multi-view representation, we first evaluate a baseline configuration using the SMILES-only representation. Specifically, only the SMILES string of a molecule is fed into the best-performing LLM model. This approach relies solely on the LLM's general understanding of a molecule and does not prompt the model to reason through the contextual diversity provided by a multi-view approach. 

\begin{figure} [!ht]
    \centering
       \includegraphics[width=0.485\textwidth]{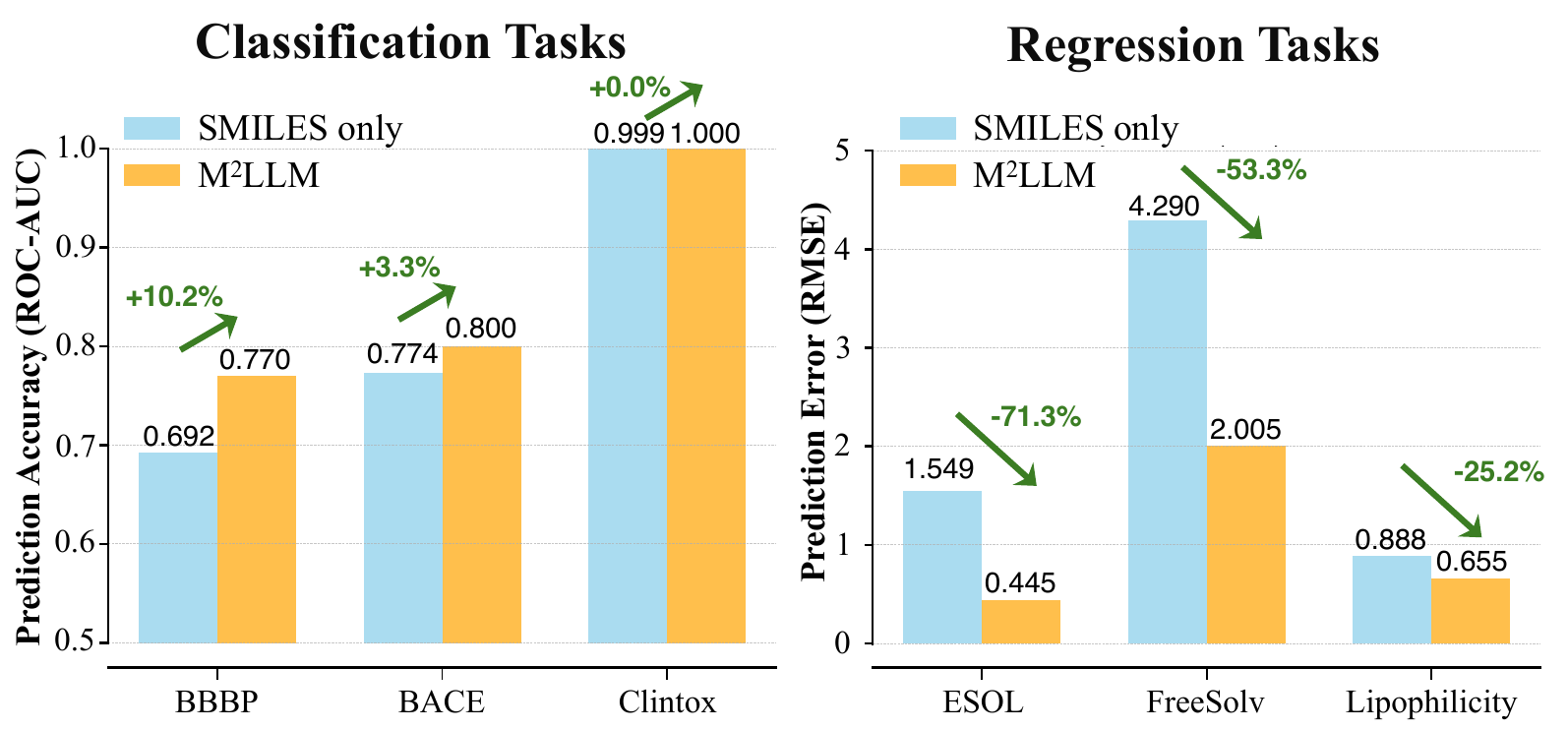}\\
       \caption{Comparison of the performance between $\text{M}^{2}$LLM and the SMILES-only representation across 10 random seeds on six datasets. } 
       \label{fig.smilesOnly}
\end{figure}

As shown in Figure \ref{fig.smilesOnly}, our proposed method consistently improved the scores across all six datasets. On the FreeSolv regression dataset, where a lower RMSE indicates better performance, we achieved a substantial reduction in prediction error, decreasing it from 4.29 to 2.01, representing a 53.3\% improvement. This improvement underscores the effectiveness of our framework, particularly in molecular property regression tasks. Similarly, for other regression and classification tasks, such as ESOL and BBBP, our method demonstrated a measurable improvement compared to the SMILES-only baseline. This trend is also consistently observed for all other LLM backbone models as illustrated in Appendix A.4.%\ref{appendix.smilesOnly}. 

In the case of the ClinTox dataset, the SMILES-only baseline achieves nearly perfect results, with our framework offering a marginal improvement. This suggests that in tasks where the LLM already possesses sufficient understanding of the molecular domain through SMILES-based encoding alone, the additional views provide less pronounced benefits. This observation, nonetheless, further reinforces the strength of LLMs as text encoders for molecular property prediction.

Overall, these results highlight the key contribution of $\text{M}^{2}$LLM: the integration of complementary molecular views consistently enhances predictive performance compared to a single-view SMILES-only approach. By dynamically incorporating diverse representations, $\text{M}^{2}$LLM captures richer molecular features and demonstrates superior adaptability across tasks of varying complexity, firmly establishing itself as a state-of-the-art framework for molecular property prediction.

%% file: Main_Text/5.Conclusion.tex
\section{Conclusion}
In this paper, we introduce $\text{M}^{2}$LLM, a multi-view learning framework that harnesses the capabilities of LLMs to generate rich molecular representations, enabling state-of-the-art performance in molecular property prediction. By utilizing the strong reasoning capabilities, extensive pre-trained knowledge, and powerful encoding abilities of LLMs, the framework delivers exceptional results across several benchmark tasks. Unlike methods that rely solely on SMILES as input, $\text{M}^{2}$LLM dynamically integrates multiple views to capture complex molecular features, enabling the learned representations to generalize effectively across diverse classification and regression tasks. These results underscore the transformative potential of $\text{M}^{2}$LLM in advancing molecular property prediction, offering a scalable and versatile solution for a wide range of applications in molecular science and beyond.

%% file: Main_Text/Appendix.tex
\clearpage
\appendix
\onecolumn

\section{Appendix}
\subsection{Dataset Description}
\label{Adx:datasets}
\begin{table}[H]
\centering

\begin{tabular}{c|c|c}
\toprule
\textbf{Dataset} & \textbf{Description} & \textbf{\# Molecules}\\ 
\hline
BBBP &Blood-brain barrier penetration dataset& 1631/204/204\\ 
\hline
BACE & Binding affinity data for human $\beta$-secretase 1 inhibitors &1209/151/152\\ 
\hline
ClinTox &Clinical trial toxicity outcomes for drugs& 1182/148/148\\ 
\hline
HIV &Molecular activity against HIV replication&32901/4113/4113 \\ 
\hline
SIDER &Drug side effect information&1141/143/143 \\ 
\hline
ESOL & Water solubility dataset & 902/113/113 \\ 
\hline
FreeSolv & Hydration free energy of small molecules in water &513/64/65\\ 
\hline
Lipophilicity & Octanol/water distribution coefficient of molecules & 3360/420/420\\ 
\bottomrule
\end{tabular}
\caption{MoleculeNet Benchmark datasets. Datasets utilized in our experiments, detailing their descriptions, the number of molecules (train/validation/test splits).}
\label{tab:datasets}
\end{table}

\subsection{Evaluation of Different Backbone Models}
\label{diff_backbone}

\begin{figure*} [!ht]
    \centering
       \includegraphics[width=0.8\textwidth]{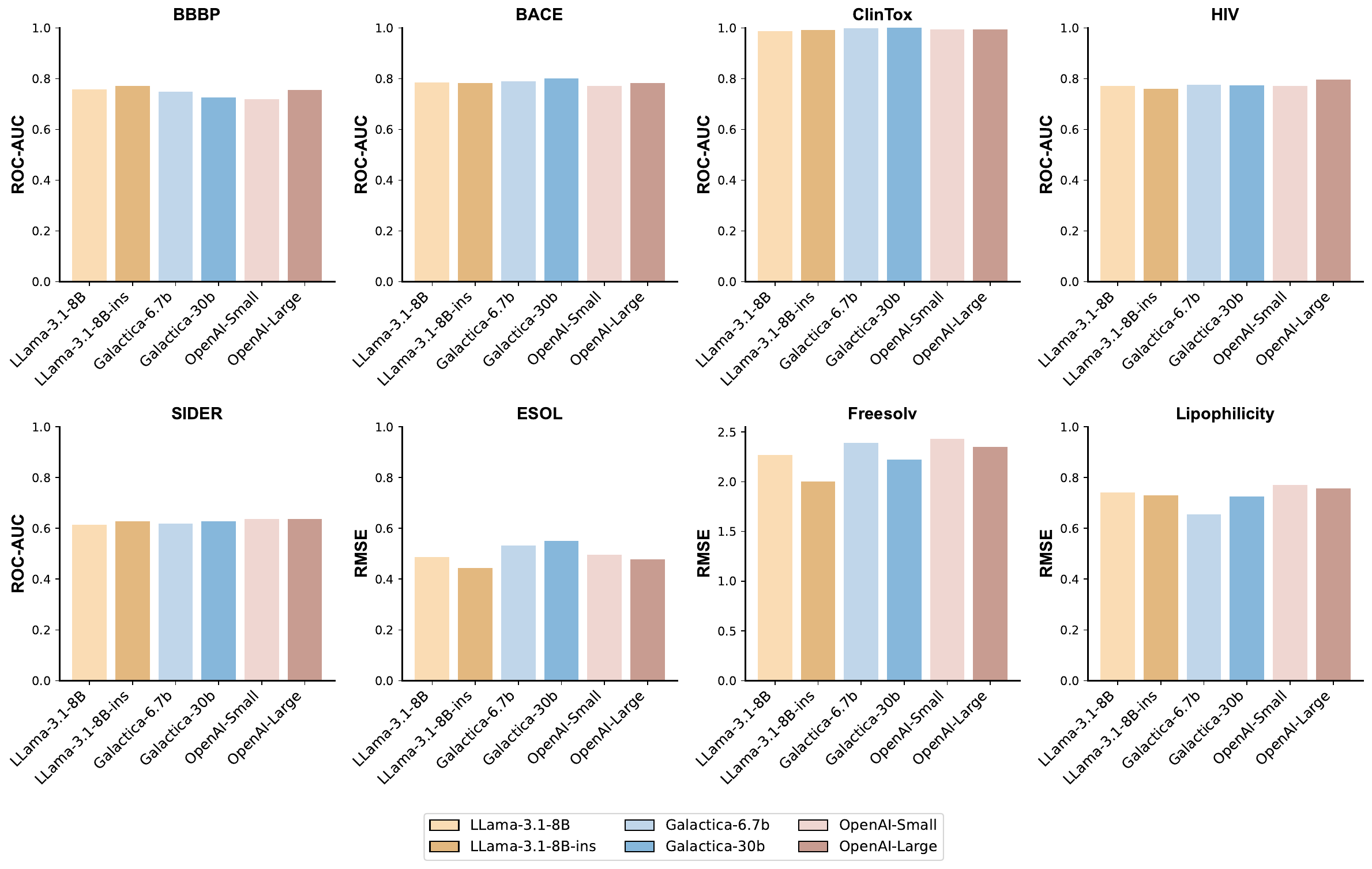}\\
       \caption{Evaluation of Different Backbone Models on 5 classification datasets and 3 regression datasets.In the classification datasets, all models achieve high ROC-AUC scores with minimal differences, indicating stable performance of our proposed framework across various tasks. For the regression datasets, our proposed framework, especially with Llama-3.1-8B-instruct, excels with the lowest RMSE scores, demonstrating its strong ability to generalize and make accurate predictions even in more challenging tasks.} 
       \label{fig.dif_backbone}
\end{figure*}

%Figure \ref{fig.dif_backbone} illustrates the evaluation of $M^{2}$LLM with different LLM backbones. In the classification datasets, all models achieve high ROC-AUC scores with minimal differences, indicating stable performance of our proposed framework across various tasks. For the regression datasets (ESOL, FreeSolv, Lipophilicity), Llama-3.1-8B-ins achieves the lowest RMSE, reinforcing its strong generalization ability. OpenAI-Large, on the other hand, tends to perform less favorably in these tasks, particularly for FreeSolv. Our proposed framework, especially with Llama-3.1-8B-instruct, excels with the lowest RMSE scores, demonstrating its strong ability to generalize and make accurate predictions even in more challenging tasks.

\newpage
\onecolumn
\subsection{Full Results on 27 tasks for the SIDER Dataset}
\label{appendix.sider}
\begin{table*}[!ht]
\resizebox{0.95\textwidth}{!}{%
\begin{tabular}{c|cc|cc|cc}
\toprule
{\multirow{2}{*}{\textbf{Sider (↑)}}} & \multicolumn{2}{c|}{\textbf{llama-3.1}} & \multicolumn{2}{c|}{\textbf{Galactica Series}} & \multicolumn{2}{c}{\textbf{OpenAI}} \\ 
\cmidrule(lr){2-3} \cmidrule(lr){4-5} \cmidrule(lr){6-7}
 & \textbf{llama-3.1-8B} & \textbf{llama-3.1-8B-instruct} & \textbf{galactica-6.7b} & \textbf{galactica-30b} & \textbf{Small} & \textbf{Large} \\ 
\midrule
cardiac disorders & 64.70 ± 0.04 & 64.78 ± 0.04 & 62.50 ± 0.03 & 63.63 ± 0.03 & 62.07 ± 0.02 & 62.96 ± 0.03 \\ 
neoplasms benign, malignant and unspecified (incl cysts and polyps) & 55.97 ± 0.03 & 57.06 ± 0.03 & 59.96 ± 0.04 & 60.35 ± 0.03 & 61.82 ± 0.02 & 63.19 ± 0.02 \\ 
hepatobiliary disorders & 57.09 ± 0.03 & 62.02 ± 0.03 & 63.32 ± 0.03 & 63.41 ± 0.03 & 60.39 ± 0.02 & 58.27 ± 0.02 \\ 
metabolism and nutrition disorders & 54.61 ± 0.02 & 55.36 ± 0.03 & 55.25 ± 0.02 & 51.85 ± 0.02 & 57.97 ± 0.02 & 54.66 ± 0.02 \\ 
product issues & 67.43 ± 0.14 & 66.85 ± 0.12 & 60.79 ± 0.11 & 66.60 ± 0.11 & 63.00 ± 0.06 & 73.50 ± 0.06 \\ 
eye disorders & 58.26 ± 0.03 & 59.14 ± 0.02 & 58.00 ± 0.03 & 59.27 ± 0.03 & 66.33 ± 0.03 & 69.47 ± 0.03 \\ 
investigations & 63.80 ± 0.06 & 66.57 ± 0.06 & 62.53 ± 0.04 & 60.74 ± 0.04 & 62.53 ± 0.04 & 75.76 ± 0.04 \\ 
musculoskeletal and connective tissue disorders & 60.42 ± 0.03 & 64.86 ± 0.05 & 63.13 ± 0.03 & 68.59 ± 0.03 & 64.24 ± 0.03 & 68.94 ± 0.02 \\ 
gastrointestinal disorders & 65.68 ± 0.04 & 66.74 ± 0.04 & 67.95 ± 0.05 & 71.16 ± 0.03 & 71.81 ± 0.04 & 61.50 ± 0.04 \\ 
social circumstances & 59.24 ± 0.03 & 56.07 ± 0.03 & 53.14 ± 0.04 & 53.34 ± 0.04 & 55.29 ± 0.04 & 59.18 ± 0.03 \\ 
immune system disorders & 62.75 ± 0.02 & 62.84 ± 0.03 & 58.61 ± 0.02 & 61.73 ± 0.03 & 63.03 ± 0.03 & 61.43 ± 0.03 \\ 
reproductive system and breast disorders & 66.16 ± 0.03 & 67.11 ± 0.02 & 64.65 ± 0.04 & 67.45 ± 0.03 & 73.53 ± 0.03 & 70.72 ± 0.03 \\ 
general disorders and administration site conditions & 57.87 ± 0.07 & 59.72 ± 0.08 & 72.48 ± 0.06 & 65.98 ± 0.04 & 55.14 ± 0.04 & 71.21 ± 0.04 \\ 
endocrine disorders & 60.60 ± 0.03 & 59.23 ± 0.03 & 68.71 ± 0.03 & 68.96 ± 0.03 & 71.79 ± 0.02 & 56.86 ± 0.04\\ 
surgical and medical procedures & 56.95 ± 0.06 & 56.15 ± 0.07 & 58.51 ± 0.05 & 63.14 ± 0.04 & 53.76 ± 0.03 & 53.73 ± 0.04 \\ 
vascular disorders & 58.03 ± 0.04 & 56.59 ± 0.03 & 59.23 ± 0.04 & 56.34 ± 0.03 & 57.92 ± 0.03 & 60.71 ± 0.03 \\ 
blood and lymphatic system disorders & 66.64 ± 0.04 & 76.29 ± 0.04 & 73.04 ± 0.03 & 72.33 ± 0.03 & 73.40 ± 0.02 & 72.13 ± 0.03 \\ 
skin and subcutaneous tissue disorders & 77.53 ± 0.08 & 76.68 ± 0.09 & 61.90 ± 0.09 & 58.69 ± 0.07 & 58.59 ± 0.04 & 59.83 ± 0.06 \\ 
congenital, familial and genetic disorders & 59.01 ± 0.04 & 61.69 ± 0.03 & 59.19 ± 0.03 & 62.88 ± 0.04 & 64.68 ± 0.03 & 65.23 ± 0.03 \\ 
infections and infestations & 63.22 ± 0.02 & 63.97 ± 0.03 & 62.97 ± 0.02 & 63.96 ± 0.02 & 63.20 ± 0.03 & 61.58 ± 0.03 \\ 
respiratory, thoracic and mediastinal disorders & 64.67 ± 0.03 & 64.95 ± 0.04 & 64.00 ± 0.03 & 62.86 ± 0.03 & 59.60 ± 0.03 & 63.94 ± 0.03 \\ 
psychiatric disorders & 58.99 ± 0.03 & 55.61 ± 0.03 & 58.84 ± 0.03 & 56.48 ± 0.03 & 63.94 ± 0.03 & 67.14 ± 0.03 \\ 
renal and urinary disorders & 59.53 ± 0.03 & 62.42 ± 0.02 & 64.17 ± 0.03 & 60.83 ± 0.03 & 76.31 ± 0.03 & 66.75 ± 0.03 \\ 
pregnancy, puerperium and perinatal conditions & 64.33 ± 0.06 & 67.32 ± 0.07 & 56.87 ± 0.06 & 65.77 ± 0.05 & 61.29 ± 0.03 & 54.64 ± 0.05 \\ 
ear and labyrinth disorders & 51.90 ± 0.02 & 54.79 ± 0.02 & 54.12 ± 0.02 & 53.64 ± 0.03 & 58.36 ± 0.02 & 53.42 ± 0.02 \\ 
nervous system disorders & 66.43 ± 0.06 & 70.90 ± 0.05 & 69.05 ± 0.05 & 70.94 ± 0.04 & 78.07 ± 0.05 & 70.63 ± 0.05 \\ 
injury, poisoning and procedural complications & 55.56 ± 0.03 & 57.37 ± 0.03 & 57.57 ± 0.03 & 63.33 ± 0.02 & 61.97 ± 0.03 & 62.49 ± 0.02 \\ 
\midrule
Average Performance & 61.38 ± 0.04 & 62.71 ± 0.04 & 61.87 ± 0.04 & 62.75 ± 0.04 & 63.70 ± 0.03 & 63.70 ± 0.03 \\ 
\bottomrule
\end{tabular}}
\caption{27 Tasks Performance Comparison for the SIDER Dataset.}
\end{table*}

\subsection{SMILES-only Representation Comparison on 6 Datasets}
\label{appendix.smilesOnly}
\begin{table}[!htbp]
    \centering
    \begin{tabular}{c|cc|cc|cc}
    \toprule
    \multirow{2}{*}{\textbf{Classification}} & \multicolumn{2}{c|}{\textbf{BBBP (↑)}} & \multicolumn{2}{c|}{\textbf{Bace (↑)}} & \multicolumn{2}{c}{\textbf{Clintox (↑)}} \\ \cmidrule(lr){2-7}
    & \textbf{Smiles\_only} & \textbf{$M^{2}$LLM} & \textbf{Smiles\_only} & \textbf{$M^{2}$LLM} & \textbf{Smiles\_only} & \textbf{$M^{2}$LLM} \\
    \midrule
    llama-3.1-8B & 67.73 ± 1.13 & 75.80 ± 0.95 & 75.16 ± 1.46 & 77.80 ± 2.97 & 99.68 ± 0.09 & 98.67 ± 2.11 \\ 
    llama-3.1-8B-instruct &69.22 ± 0.84 & 77.04 ± 1.00 & 75.32 ± 1.47 & 77.43 ± 1.89 & 99.33 ± 0.12 & 99.11 ± 0.42 \\ 
    \midrule
    galactica-6.7b & 69.17 ± 1.12 & 74.94 ± 0.74 & 74.56 ± 1.86 & 78.88 ± 2.16 & 100.00 ± 0.00 & 99.93 ± 0.05 \\ 
    galactica-30b & 68.57 ± 1.63 & 72.57 ± 1.32 & 77.37 ± 1.18 & 79.98 ± 2.67 & 99.99 ± 0.02 & 100.00 ± 0.00 \\ 
    \midrule
    text-embedding-3-small & 71.54 ± 3.74 & 71.89 ± 1.36 & 75.96 ± 0.92 & 77.24 ± 1.65 & 99.64 ± 0.06 & 99.49 ± 0.07 \\ 
    text-embedding-3-large & 71.65 ± 3.08 & 75.50 ± 1.26 & 77.31 ± 1.77 & 78.18 ± 0.96 & 99.86 ± 0.00 & 99.38 ± 0.09 \\ 
    \bottomrule
    \end{tabular}
\caption{Classification results for BBBP, Bace, and Clintox datasets.}
\label{table_sonlyCls}
\end{table}

\begin{table}[!ht]
    \centering
    \begin{tabular}{c|cc|cc|cc}
    \toprule
    \multirow{2}{*}{\textbf{Regression}} & \multicolumn{2}{c|}{\textbf{ESOL (↓)}} & \multicolumn{2}{c|}{\textbf{FreeSolv (↓)}} & \multicolumn{2}{c}{\textbf{Lipophilicity (↓)}} \\ \cmidrule(lr){2-7}
    & \textbf{Smiles\_only} & \textbf{$M^{2}$LLM} & \textbf{Smiles\_only} & \textbf{$M^{2}$LLM} & \textbf{Smiles\_only} & \textbf{$M^{2}$LLM} \\
    \midrule
    llama-3.1-8B & 1.53 ± 0.03 & 0.49 ± 0.05 & 4.01 ± 0.73 & 2.27 ± 0.84 & 0.93 ± 0.02 & 0.74 ± 0.05 \\ 
    llama-3.1-8B-instruct & 1.55 ± 0.03 & 0.44 ± 0.01 & 4.29 ± 0.93 & 2.01 ± 0.37 & 0.93 ± 0.02 & 0.73 ± 0.03 \\ 
    \midrule
    galactica-6.7b & 1.69 ± 0.07 & 0.53 ± 0.25 & 4.58 ± 1.62 & 2.39 ± 1.39 & 0.89 ± 0.02 & 0.66 ± 0.02 \\ 
    galactica-30b & 1.39 ± 0.02 & 0.55 ± 0.24 & 7.79 ± 1.34 & 2.22 ± 0.74 & 0.87 ± 0.03 & 0.73 ± 0.03 \\ 
    \midrule
    text-embedding-3-small & 3.29 ± 0.09 & 0.50 ± 0.08 & 3.65 ± 0.23 & 2.43 ± 0.60 & 0.91 ± 0.01 & 0.77 ± 0.01 \\ 
    text-embedding-3-large & 2.11 ± 0.10 & 0.48 ± 0.02 & 3.73 ± 0.83 & 2.35 ± 0.63 & 0.87 ± 0.01 & 0.79 ± 0.02 \\ 
    \bottomrule
    \end{tabular}
\caption{Regression results for ESOL, FreeSolv, and Lipophilicity datasets.}
\label{table_sonlyReg}
\end{table}

%% file: 0.Main.bbl
\begin{thebibliography}{}

\bibitem[\protect\citeauthoryear{Breiman}{2001}]{breiman2001random}
Leo Breiman.
\newblock Random forests.
\newblock {\em Machine learning}, 45:5--32, 2001.

\bibitem[\protect\citeauthoryear{Bu \bgroup \em et al.\egroup }{2024}]{bu2024improving}
Weixin Bu, Xiaofeng Cao, Yizhen Zheng, and Shirui Pan.
\newblock Improving augmentation consistency for graph contrastive learning.
\newblock {\em Pattern Recognition}, 148:110182, 2024.

\bibitem[\protect\citeauthoryear{Delaney}{2004}]{delaney2004esol}
John~S Delaney.
\newblock Esol: estimating aqueous solubility directly from molecular structure.
\newblock {\em Journal of chemical information and computer sciences}, 44(3):1000--1005, 2004.

\bibitem[\protect\citeauthoryear{Drews}{2000}]{drews2000drug}
Jurgen Drews.
\newblock Drug discovery: a historical perspective.
\newblock {\em science}, 287(5460):1960--1964, 2000.

\bibitem[\protect\citeauthoryear{Du \bgroup \em et al.\egroup }{2024}]{du2024mmgnn}
Wenjie Du, Shuai Zhang, Jun~Xia Di~Wu, Ziyuan Zhao, Junfeng Fang, and Yang Wang.
\newblock Mmgnn: A molecular merged graph neural network for explainable solvation free energy prediction.
\newblock In {\em Proceedings of the Thirty-Third International Joint Conference on Artificial Intelligence}, pages 5808--5816, 2024.

\bibitem[\protect\citeauthoryear{Dubey \bgroup \em et al.\egroup }{2024}]{dubey2024llama}
Abhimanyu Dubey, Abhinav Jauhri, et~al.
\newblock The llama 3 herd of models.
\newblock {\em arXiv preprint arXiv:2407.21783}, 2024.

\bibitem[\protect\citeauthoryear{Fabian \bgroup \em et al.\egroup }{2020}]{fabian2020molecular}
Benedek Fabian, Thomas Edlich, et~al.
\newblock Molecular representation learning with language models and domain-relevant auxiliary tasks.
\newblock {\em arXiv preprint arXiv:2011.13230}, 2020.

\bibitem[\protect\citeauthoryear{Gayvert \bgroup \em et al.\egroup }{2016}]{gayvert2016data}
Kaitlyn~M Gayvert, Neel~S Madhukar, and Olivier Elemento.
\newblock A data-driven approach to predicting successes and failures of clinical trials.
\newblock {\em Cell chemical biology}, 23(10):1294--1301, 2016.

\bibitem[\protect\citeauthoryear{Harding \bgroup \em et al.\egroup }{2024}]{harding2024iuphar}
Simon~D Harding, Jane~F Armstrong, et~al.
\newblock The iuphar/bps guide to pharmacology in 2024.
\newblock {\em Nucleic Acids Research}, 52(D1):D1438--D1449, 2024.

\bibitem[\protect\citeauthoryear{Hu \bgroup \em et al.\egroup }{2019}]{hu2019strategies}
Weihua Hu, Bowen Liu, et~al.
\newblock Strategies for pre-training graph neural networks.
\newblock {\em arXiv preprint arXiv:1905.12265}, 2019.

\bibitem[\protect\citeauthoryear{Jeon and Kim}{2019}]{jeon2019fp2vec}
Woosung Jeon and Dongsup Kim.
\newblock Fp2vec: a new molecular featurizer for learning molecular properties.
\newblock {\em Bioinformatics}, 35(23):4979--4985, 2019.

\bibitem[\protect\citeauthoryear{Kenton and Toutanova}{2019}]{kenton2019bert}
Jacob Devlin Ming-Wei~Chang Kenton and Lee~Kristina Toutanova.
\newblock Bert: Pre-training of deep bidirectional transformers for language understanding.
\newblock In {\em Proceedings of naacL-HLT}, volume~1. Minneapolis, Minnesota, 2019.

\bibitem[\protect\citeauthoryear{Koh \bgroup \em et al.\egroup }{2024}]{koh2024physicochemical}
Huan~Yee Koh, Anh~TN Nguyen, Shirui Pan, Lauren~T May, and Geoffrey~I Webb.
\newblock Physicochemical graph neural network for learning protein--ligand interaction fingerprints from sequence data.
\newblock {\em Nature Machine Intelligence}, pages 1--15, 2024.

\bibitem[\protect\citeauthoryear{Kojima \bgroup \em et al.\egroup }{2022}]{kojima2022large}
Takeshi Kojima, Shixiang~Shane Gu, et~al.
\newblock Large language models are zero-shot reasoners.
\newblock {\em Advances in neural information processing systems}, 35:22199--22213, 2022.

\bibitem[\protect\citeauthoryear{Kuhn \bgroup \em et al.\egroup }{2016}]{kuhn2016sider}
Michael Kuhn, Ivica Letunic, Lars~Juhl Jensen, and Peer Bork.
\newblock The sider database of drugs and side effects.
\newblock {\em Nucleic acids research}, 44(D1):D1075--D1079, 2016.

\bibitem[\protect\citeauthoryear{Landrum and Riniker}{2024}]{landrum2024combining}
Gregory~A Landrum and Sereina Riniker.
\newblock Combining ic50 or k i values from different sources is a source of significant noise.
\newblock {\em Journal of Chemical Information and Modeling}, 64(5):1560--1567, 2024.

\bibitem[\protect\citeauthoryear{Liu \bgroup \em et al.\egroup }{2022}]{liu2022pre}
Shengchao Liu, Hanchen Wang, et~al.
\newblock Pre-training molecular graph representation with 3d geometry.
\newblock In {\em International Conference on Learning Representations}, 2022.

\bibitem[\protect\citeauthoryear{Luo \bgroup \em et al.\egroup }{2024}]{luo2024learning}
Yizhen Luo, Kai Yang, et~al.
\newblock Learning multi-view molecular representations with structured and unstructured knowledge.
\newblock In {\em Proceedings of the 30th ACM SIGKDD Conference on Knowledge Discovery and Data Mining}, pages 2082--2093, 2024.

\bibitem[\protect\citeauthoryear{Martins \bgroup \em et al.\egroup }{2012}]{martins2012bayesian}
Ines~Filipa Martins, Ana~L Teixeira, et~al.
\newblock A bayesian approach to in silico blood-brain barrier penetration modeling.
\newblock {\em Journal of chemical information and modeling}, 52(6):1686--1697, 2012.

\bibitem[\protect\citeauthoryear{Medsker \bgroup \em et al.\egroup }{2001}]{medsker2001recurrent}
Larry~R Medsker, Lakhmi Jain, et~al.
\newblock Recurrent neural networks.
\newblock {\em Design and Applications}, 5(64-67):2, 2001.

\bibitem[\protect\citeauthoryear{Mirza \bgroup \em et al.\egroup }{2024}]{mirza2024large}
Adrian Mirza, Nawaf Alampara, et~al.
\newblock Are large language models superhuman chemists?
\newblock {\em arXiv preprint arXiv:2404.01475}, 2024.

\bibitem[\protect\citeauthoryear{Mobley and Guthrie}{2014}]{mobley2014freesolv}
David~L Mobley and J~Peter Guthrie.
\newblock Freesolv: a database of experimental and calculated hydration free energies, with input files.
\newblock {\em Journal of computer-aided molecular design}, 28:711--720, 2014.

\bibitem[\protect\citeauthoryear{OpenAI \bgroup \em et al.\egroup }{2023}]{achiam2023gpt}
Josh OpenAI, Achiam, Steven Adler, et~al.
\newblock Gpt-4 technical report.
\newblock {\em arXiv preprint arXiv:2303.08774}, 2023.

\bibitem[\protect\citeauthoryear{{OpenAI}}{2024}]{openai2024embedding}
{OpenAI}.
\newblock New embedding models and api updates, 2024.

\bibitem[\protect\citeauthoryear{Rogers and Hahn}{2010}]{rogers2010extended}
David Rogers and Mathew Hahn.
\newblock Extended-connectivity fingerprints.
\newblock {\em Journal of chemical information and modeling}, 50(5):742--754, 2010.

\bibitem[\protect\citeauthoryear{Rollins \bgroup \em et al.\egroup }{2024}]{rollins2024molprop}
Zachary~A Rollins, Alan~C Cheng, and Essam Metwally.
\newblock Molprop: Molecular property prediction with multimodal language and graph fusion.
\newblock {\em Journal of Cheminformatics}, 16(1):56, 2024.

\bibitem[\protect\citeauthoryear{Rong \bgroup \em et al.\egroup }{2020}]{rong2020self}
Yu~Rong, Yatao Bian, et~al.
\newblock Self-supervised graph transformer on large-scale molecular data.
\newblock {\em Advances in neural information processing systems}, 33:12559--12571, 2020.

\bibitem[\protect\citeauthoryear{Ross \bgroup \em et al.\egroup }{2022}]{ross2022large}
Jerret Ross, Brian Belgodere, et~al.
\newblock Large-scale chemical language representations capture molecular structure and properties.
\newblock {\em Nature Machine Intelligence}, 4(12):1256--1264, 2022.

\bibitem[\protect\citeauthoryear{Sadeghi \bgroup \em et al.\egroup }{2024}]{sadeghi2024comparative}
Shaghayegh Sadeghi, Alan Bui, Ali Forooghi, Jianguo Lu, and Alioune Ngom.
\newblock Comparative analysis of llama and chatgpt embeddings for molecule embedding.
\newblock {\em arXiv preprint arXiv:2402.00024}, 2024.

\bibitem[\protect\citeauthoryear{Shirasuna \bgroup \em et al.\egroup }{2024}]{shirasuna2024multi}
Victor~Yukio Shirasuna, Eduardo Soares, et~al.
\newblock A multi-view mixture-of-experts based on language and graphs for molecular properties prediction.
\newblock In {\em ICML 2024 AI for Science Workshop}, 2024.

\bibitem[\protect\citeauthoryear{St{\"a}rk \bgroup \em et al.\egroup }{2022}]{stark20223d}
Hannes St{\"a}rk, Dominique Beaini, Gabriele Corso, Prudencio Tossou, Christian Dallago, Stephan G{\"u}nnemann, and Pietro Li{\`o}.
\newblock 3d infomax improves gnns for molecular property prediction.
\newblock In {\em International Conference on Machine Learning}, pages 20479--20502. PMLR, 2022.

\bibitem[\protect\citeauthoryear{Subramanian \bgroup \em et al.\egroup }{2016}]{subramanian2016computational}
Govindan Subramanian, Bharath Ramsundar, Vijay Pande, and Rajiah~Aldrin Denny.
\newblock Computational modeling of $\beta$-secretase 1 (bace-1) inhibitors using ligand based approaches.
\newblock {\em Journal of chemical information and modeling}, 56(10):1936--1949, 2016.

\bibitem[\protect\citeauthoryear{Taylor \bgroup \em et al.\egroup }{2022}]{taylor2022galactica}
Ross Taylor, Marcin Kardas, Guillem Cucurull, Thomas Scialom, Anthony Hartshorn, Elvis Saravia, Andrew Poulton, Viktor Kerkez, and Robert Stojnic.
\newblock Galactica: A large language model for science.
\newblock {\em arXiv preprint arXiv:2211.09085}, 2022.

\bibitem[\protect\citeauthoryear{Wang \bgroup \em et al.\egroup }{2019}]{wang2019smiles}
Sheng Wang, Yuzhi Guo, et~al.
\newblock Smiles-bert: large scale unsupervised pre-training for molecular property prediction.
\newblock In {\em Proceedings of the 10th ACM international conference on bioinformatics, computational biology and health informatics}, pages 429--436, 2019.

\bibitem[\protect\citeauthoryear{Wang \bgroup \em et al.\egroup }{2022}]{wang2022molecular}
Yuyang Wang, Jianren Wang, Zhonglin Cao, and Amir Barati~Farimani.
\newblock Molecular contrastive learning of representations via graph neural networks.
\newblock {\em Nature Machine Intelligence}, 4(3):279--287, 2022.

\bibitem[\protect\citeauthoryear{Wang \bgroup \em et al.\egroup }{2024a}]{wang2024goodat}
Luzhi Wang, Dongxiao He, He~Zhang, Yixin Liu, Wenjie Wang, Shirui Pan, Di~Jin, and Tat-Seng Chua.
\newblock Goodat: towards test-time graph out-of-distribution detection.
\newblock In {\em Proceedings of the AAAI Conference on Artificial Intelligence}, volume~38, pages 15537--15545, 2024.

\bibitem[\protect\citeauthoryear{Wang \bgroup \em et al.\egroup }{2024b}]{wang2024contrastive}
Luzhi Wang, Yizhen Zheng, Di~Jin, Fuyi Li, Yongliang Qiao, and Shirui Pan.
\newblock Contrastive graph similarity networks.
\newblock {\em ACM Transactions on the Web}, 18(2):1--20, 2024.

\bibitem[\protect\citeauthoryear{Weininger}{1988}]{weininger1988smiles}
David Weininger.
\newblock Smiles, a chemical language and information system. 1. introduction to methodology and encoding rules.
\newblock {\em Journal of chemical information and computer sciences}, 28(1):31--36, 1988.

\bibitem[\protect\citeauthoryear{Wu \bgroup \em et al.\egroup }{2018}]{wu2018moleculenet}
Zhenqin Wu, Bharath Ramsundar, Evan~N Feinberg, Joseph Gomes, Caleb Geniesse, Aneesh~S Pappu, Karl Leswing, and Vijay Pande.
\newblock Moleculenet: a benchmark for molecular machine learning.
\newblock {\em Chemical science}, 9(2):513--530, 2018.

\bibitem[\protect\citeauthoryear{Wu \bgroup \em et al.\egroup }{2024}]{wu2024graph}
Man Wu, Xin Zheng, Qin Zhang, Xiao Shen, Xiong Luo, Xingquan Zhu, and Shirui Pan.
\newblock Graph learning under distribution shifts: A comprehensive survey on domain adaptation, out-of-distribution, and continual learning.
\newblock {\em arXiv preprint arXiv:2402.16374}, 2024.

\bibitem[\protect\citeauthoryear{Xia \bgroup \em et al.\egroup }{2022}]{xia2022mole}
Jun Xia, Chengshuai Zhao, Bozhen Hu, Zhangyang Gao, Cheng Tan, Yue Liu, Siyuan Li, and Stan~Z Li.
\newblock Mole-bert: Rethinking pre-training graph neural networks for molecules.
\newblock In {\em The Eleventh International Conference on Learning Representations}, 2022.

\bibitem[\protect\citeauthoryear{Yang \bgroup \em et al.\egroup }{2019}]{yang2019analyzing}
Kevin Yang, Kyle Swanson, Wengong Jin, et~al.
\newblock Analyzing learned molecular representations for property prediction.
\newblock {\em Journal of chemical information and modeling}, 59(8):3370--3388, 2019.

\bibitem[\protect\citeauthoryear{You \bgroup \em et al.\egroup }{2020}]{you2020graph}
Yuning You, Tianlong Chen, Yongduo Sui, Ting Chen, Zhangyang Wang, and Yang Shen.
\newblock Graph contrastive learning with augmentations.
\newblock {\em Advances in neural information processing systems}, 33:5812--5823, 2020.

\bibitem[\protect\citeauthoryear{Yu \bgroup \em et al.\egroup }{2024}]{yu2024kernel}
Jiajun Yu, Zhihao Wu, Jinyu Cai, Adele~Lu Jia, and Jicong Fan.
\newblock Kernel readout for graph neural networks.
\newblock In {\em Proceedings of the Thirty-Third International Joint Conference on Artificial Intelligence, IJCAI-24}, pages 2505--2514, 2024.

\bibitem[\protect\citeauthoryear{Yu \bgroup \em et al.\egroup }{2025}]{yu2025collaborative}
Jiajun Yu, Yizhen Zheng, Huan~Yee Koh, Shirui Pan, Tianyue Wang, and Haishuai Wang.
\newblock Collaborative expert llms guided multi-objective molecular optimization.
\newblock {\em arXiv preprint arXiv:2503.03503}, 2025.

\bibitem[\protect\citeauthoryear{Zhang \bgroup \em et al.\egroup }{2019}]{ZhangMHLLL19}
He~Zhang, Hanlin Mo, You Hao, Qi~Li, Shirui Li, and Hua Li.
\newblock Fast and efficient calculations of structural invariants of chirality.
\newblock {\em Pattern Recognit. Lett.}, 128:270--277, 2019.

\bibitem[\protect\citeauthoryear{Zhang \bgroup \em et al.\egroup }{2025}]{zhang2025dynamic}
He~Zhang, Bang Wu, Xiangwen Yang, Xingliang Yuan, Xiaoning Liu, and Xun Yi.
\newblock Dynamic graph unlearning: A general and efficient post-processing method via gradient transformation.
\newblock In {\em Proceedings of the ACM on Web Conference 2025}, pages 931--944, 2025.

\bibitem[\protect\citeauthoryear{Zheng \bgroup \em et al.\egroup }{2024a}]{zhengonline}
Xin Zheng, Dongjin Song, Qingsong Wen, Bo~Du, and Shirui Pan.
\newblock Online gnn evaluation under test-time graph distribution shifts.
\newblock In {\em The Twelfth International Conference on Learning Representations}, 2024.

\bibitem[\protect\citeauthoryear{Zheng \bgroup \em et al.\egroup }{2024b}]{zhengcross}
Yan Zheng, Song Wu, Junyu Lin, Yazhou Ren, Jing He, Xiaorong Pu, and Lifang He.
\newblock Cross-view contrastive fusion for enhanced molecular property prediction.
\newblock {\em Proccedings of the Thirty-Third International Joint Conference on Artificial Intelligence}, 2024.

\bibitem[\protect\citeauthoryear{Zheng \bgroup \em et al.\egroup }{2024c}]{zheng2024large}
Yizhen Zheng, Huan~Yee Koh, Maddie Yang, Li~Li, Lauren~T May, Geoffrey~I Webb, Shirui Pan, and George Church.
\newblock Large language models in drug discovery and development: From disease mechanisms to clinical trials.
\newblock {\em arXiv preprint arXiv:2409.04481}, 2024.

\bibitem[\protect\citeauthoryear{Zheng \bgroup \em et al.\egroup }{2025}]{zheng2023large}
Yizhen Zheng, Huan~Yee Koh, Jiaxin Ju, Anh~TN Nguyen, Lauren~T May, Geoffrey~I Webb, and Shirui Pan.
\newblock Large language models for scientific discovery in molecular property prediction.
\newblock {\em Nature Machine Intelligence}, pages 1--11, 2025.

\bibitem[\protect\citeauthoryear{Zhou \bgroup \em et al.\egroup }{2023}]{zhou2023unimol}
Gengmo Zhou, Zhifeng Gao, et~al.
\newblock Uni-mol: A universal 3d molecular representation learning framework.
\newblock In {\em The Eleventh International Conference on Learning Representations}, 2023.

\end{thebibliography}
